\documentclass[11pt]{article}
\usepackage[final]{acl}
\usepackage{times}
\usepackage{latexsym}
\usepackage[T1]{fontenc}
\usepackage[utf8]{inputenc}
\usepackage{microtype}
\usepackage{inconsolata}
\usepackage{graphicx}
\usepackage{enumitem}  
\usepackage{amsmath}  
\usepackage{amsfonts}
\usepackage{amsthm}
\usepackage{adjustbox}
\usepackage{booktabs}
\usepackage{multirow}
\usepackage{amsmath}
\usepackage{amssymb}
\usepackage{mathtools}
\usepackage{amsthm}
\usepackage[capitalize,noabbrev]{cleveref}
\theoremstyle{plain}
\newtheorem{theorem}{Theorem}[section]

\theoremstyle{definition}

\theoremstyle{remark}

\usepackage[textsize=tiny]{todonotes}
\usepackage{adjustbox}
\usepackage[most]{tcolorbox}
\usepackage{listings}
\usepackage{xcolor}  
\usepackage{graphicx}  
\usepackage{enumitem}
\usepackage{multirow}

\title{VEM: Environment-Free Exploration for Training GUI Agent with\\ Value Environment Model}

\author{
 \textbf{Jiani Zheng\textsuperscript{1, \thanks{Work done during the internship at Microsoft.}}},
 \textbf{Lu Wang\textsuperscript{2}},
 \textbf{Fangkai Yang\textsuperscript{2}},
 \textbf{Chaoyun Zhang\textsuperscript{2}},
 \textbf{Lingrui Mei\textsuperscript{3,\footnotemark[1]}},
 \textbf{Wenjie Yin\textsuperscript{4}}, \\
 \textbf{Qingwei Lin\textsuperscript{2}},
 \textbf{Dongmei Zhang\textsuperscript{2}},
 \textbf{Saravan Rajmohan\textsuperscript{2}},
 \textbf{Qi Zhang\textsuperscript{2}}
\\
\\
 \textsuperscript{1}Peking University,
 \textsuperscript{2}Microsoft,
\\
 \textsuperscript{3}University of the Chinese Academy of Sciences,
 \textsuperscript{4}KTH Royal Institute of Technology
\\
 \small{
   \texttt{\{wlu, fangkaiyang, chaoyunzhang\}@microsoft.com}
 }
}

\begin{document}

\maketitle

\begin{abstract}
Training Vision-Language Models (VLMs) for Graphical User Interfaces (GUI) agents via Reinforcement Learning (RL) faces critical challenges: environment-based RL requires costly interactions, while environment-free methods struggle with distribution shift and reward generalization. We propose an environment-free RL framework that decouples value estimation from policy optimization by leveraging a pretrained Value Environment Model (VEM). VEM predicts state-action values directly from offline data, distilling human-like priors about GUI interaction outcomes without requiring next-state prediction or environmental feedback. This avoids compounding errors and enhances resilience to UI changes by focusing on semantic reasoning (e.g., “Does this action advance the user’s goal?”). The framework operates in two stages: (1) pretraining VEM to estimate long-term action utilities and (2) guiding policy exploration with frozen VEM signals, enabling layout-agnostic GUI automation. Evaluated on Android-in-the-Wild benchmarks, VEM achieves state-of-the-art performance in both offline and online settings, outperforming environment-free baselines significantly and matching environment-based approaches without interaction costs. Importantly, VEM demonstrates that semantic-aware value estimation can achieve comparable performance with online-trained methods. The code is available at: \url{https://github.com/microsoft/GUI-Agent-RL}, and the full project page can be found at  \href{https://redesigned-adventure-8q68lwk.pages.github.io/}{VEM}.
\end{abstract}

\section{Introduction}
\label{sec:intro}
Vision-Language Models (VLMs) have demonstrated remarkable capabilities in tasks requiring common-sense reasoning, abstraction, and generalization, enabling their use in a wide range of applications~\cite{zhou2022learning,zhang2024vision}, including autonomous Graphical User Interfaces (GUI) agents \cite{zhang2024large,wang2024gui,nguyen2024gui}. GUI agents leverage the VLM's ability to process visual and language information for solving device control tasks that automate interactions with GUIs effectively. For example, they can decipher natural language instructions like ``\textit{Find the `Save' button and click it}'' and use visual data to locate the relevant element on the GUI. This makes them valuable in automating routine GUI operations, from simple web navigation tasks to complex software interactions and management~\cite{yan2023gpt,zhang2023you,zhang2023appagent,rawles2024androidinthewild, bai2024digirl, hong2024cogagent}. 
Despite the progress in general-purpose VLMs like GPT-4o~\cite{GPT-4o} and Gemini 1.5 Pro~\cite{Gemini_1.5_Pro}, these models face significant challenges in completing real-world GUI tasks, even combined with prompting and tool usage~\cite{xie2024osworld,zhang2024ufo,zhang2024android,bai2024digirl}. For example, they may misinterpret visual cues due to the complexity and variability of GUIs. A minor change in the layout or design of a GUI, like pop-up windows or repositioned button, can cause the model to make mistakes~\cite{zhang2024large,zhang2024ufo}. 

As such, there is an urgent need to train specialized VLMs tailored to GUI tasks, enabling GUI agents to handle a wider variety of GUI tasks with higher precision and efficiency. Reinforcement Learning (RL)~\cite{sutton2018reinforcement} is a proved-effective way to align VLMs or LLMs with desired behaviors \cite{zhai2024fine,sun2024llm}. In approaches involving \textit{environment-based RL}, models interact with environments where rewards are obtained directly from the environment~\cite{toyama2021androidenv,bai2024digirl,carta2023grounding,wang2024distrl,lai2024autowebglm}. However, these methods are limited by online environmental reward dependency which is costly or unavailable, resulting in sample inefficiency~\cite{xie2021policy,niu2022trust}. Other works simulate an environment to predict the next state and give rewards~\cite{chae2024web,gu2024your}, but suffer from the fidelity issues and have compounding errors in sequential next state predictions~\cite{guan2023leveraging,zhang2024mme,ge2024worldgpt}. On the other hand, \textit{environment-free RL} employs techniques such as offline RL~\cite{snell2022offline,hong2023zero,bai2024digirl,wang2024large} or training reward models~\cite{stiennon2020learning,ouyang2022training}, allowing fine-tuning of large models using offline data without direct interaction with an environment. However, offline RL face challenges such as distribution shift~\cite{levine2020offline} and limited exploration~\cite{prudencio2023survey}. Similarly, methods relying on learned reward models~\cite{stiennon2020learning} often struggle to generalize beyond the static reward signals present in offline data, particularly in dynamic GUI environments where visual and functional changes (e.g., UI redesigns) can invalidate pre-defined reward criteria.

\begin{figure}[t]
  \centering
  \includegraphics[width=0.95\linewidth]{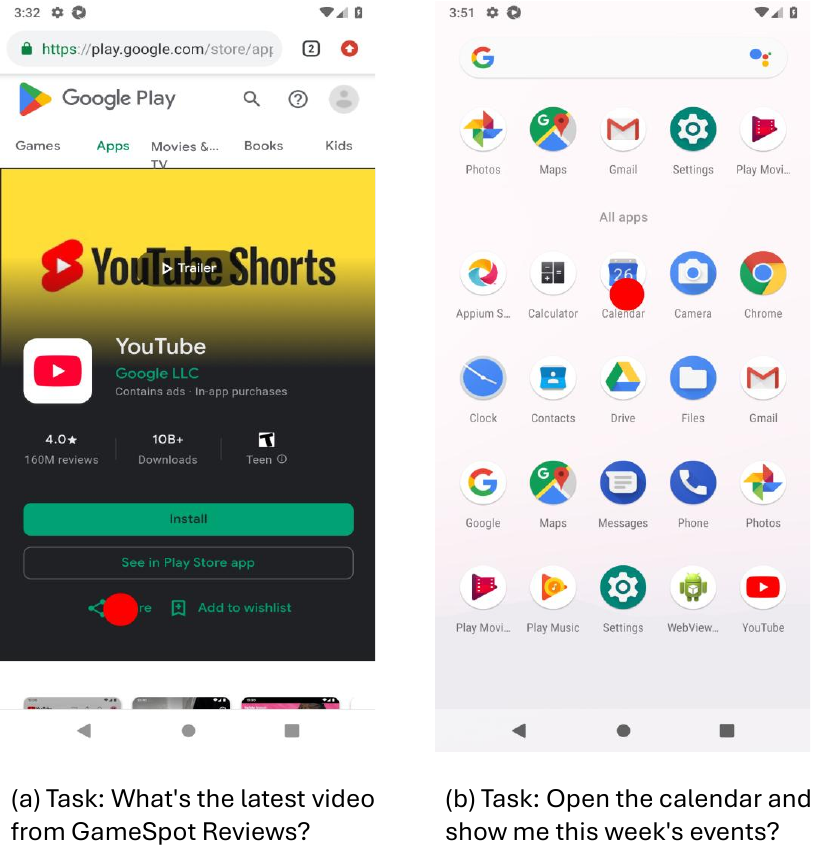}
  \caption{Two GUI tasks with the action marked as a red dot (\textcolor{red}{\raisebox{-0.25ex}{\scalebox{2}{$\bullet$}}}). Although we do not see the next state after taking the action, we can estimate the state-action value. In (a), clicking the 'share' button is unlikely to reveal reviews, implying a low state-action value. In (b), the action may open the calendar app, making it a plausible step to display the week's events, suggesting a high state-action value.}
  \label{fig:intro_case}
  \vspace{-5mm}
\end{figure}

As shown in Figure~\ref{fig:intro_case}, humans excel at estimating the long-term utility of actions (i.e., state-action values $Q(s,a)$) without explicitly seeing the next state. VLMs, trained on vast corpora of human-GUI interactions, inherently encode similar priors about action outcomes~\cite{hao2023reasoning,bai2023qwen,chen2023minigpt}. To operationalize this capability and address aforementioned challenges, we propose an environment-free RL framework that decouples value estimation from policy optimization. Unlike prior work that depends on reward models or direct environment interactions, our method leverages a pretrained \textit{value environment model}~(VEM) to approximate state-action values directly from offline data. By distilling human-like priors into a frozen VEM, our policy model bypasses the need for explicit reward engineering or error-prone next-state simulations. Additionally, the VEM’s reliance on semantic reasoning (e.g., ``Does this action advance the user’s goal?'') instead of pixel-level next-state predictions makes it resilient to superficial UI changes.

Concretely, our framework operates in two stages:
\begin{enumerate}[leftmargin=*]
    \item \textbf{Value Environment Model Pretraining:} The VEM is trained offline to predict state-action values $Q(s,a)$, capturing the long-term utility of actions in diverse GUI contexts. This avoids the compounding errors of next-state prediction by focusing on value estimation, which aligns better with VLMs’ inherent reasoning strengths.
    \item \textbf{Policy Exploration with Frozen VEM:} During policy training, the VEM provides value-guided signals to iteratively refine the policy’s action selection. By directly exploring for high-value actions that grounded in the VEM’s understanding of GUI semantics, the policy learns to generalize across unseen layouts and functionalities without online interaction.
\end{enumerate}

To evaluate our method, we conduct rigorous experiments on Android-in-the-Wild (AITW)~\cite{rawles2024androidinthewild} using dual offline/online protocols. With only 500 training trajectories (one-third of the DigiRL~\cite{bai2024digirl} dataset scale data), our approach achieves 28.0\%/21.0\% offline task success on General/Webshopping domains, outperforming environment-free counterparts by 12–28\% and matching or exceeding environment-based approaches by 3–5\%. In online deployment, we attain 42.4\% general task success, surpassing environment-free methods by 14–43\% while remaining comparable to environment-based policies (38.98\%) in procedural efficiency (7.83 vs. 7.25 average steps). Crucially, our method eliminates catastrophic failures of generic models (e.g., 0\% webshopping success in GPT-4o) through structured credit assignment, achieving 21.7\% relative improvement over the strongest baseline without environmental interaction costs. These results demonstrate that offline policy optimization can rival online-trained systems while significantly advancing environment-free paradigms.

\section{Related Works}
\subsection{Environment-Based Methods}
Environment-based RL methods train VLMs through direct interaction with GUI environments, where rewards are explicitly provided by the environment. Several frameworks like AndroidEnv~\cite{toyama2021androidenv} and DistRL~\cite{wang2024distrl} enable agents to learn through trial-and-error interactions with real-world digital interfaces. Recent works such as DigiRL~\cite{bai2024digirl} and AutoWebGLM~\cite{lai2024autowebglm} demonstrate that environment-based RL can effectively align VLMs with complex GUI navigation tasks through autonomous exploration. However, these methods face significant limitations in sample efficiency due to the high cost of environment interactions~\cite{xie2021policy,niu2022trust}, particularly in real-world applications where collecting online feedback is expensive and suffers from high latency. To address this, some approaches like WebRL~\cite{qi2024webrl} and WorldGPT~\cite{ge2024worldgpt} attempt to simulate GUI environments, but they struggle with state prediction accuracy and compounding errors in long interaction sequences~\cite{guan2023leveraging,zhang2024mme}. The fundamental challenge lies in the dynamic nature of real-world GUIs, where interface elements and layouts frequently change, making environment-dependent reward signals inherently unstable~\cite{zhang2024ufo,zhang2024android}.

\subsection{Environment-Free Methods}
Environment-free approaches bypass direct environment interaction by leveraging offline datasets or learned reward models. Offline RL methods~\cite{snell2022offline,hong2023zero} have shown promise in fine-tuning VLMs using pre-collected interaction trajectories, as demonstrated in large-scale GUI agent training by~\cite{wang2024large} and~\cite{bai2024digirl}. Alternative approaches employ reward modeling techniques~\cite{stiennon2020learning,ouyang2022training} to predict task success signals from static datasets, enabling behavior alignment without environment feedback. General-purpose VLMs like GPT-4o~\cite{GPT-4o} represent a distinct category of environment-free methods, leveraging their pre-trained visual-language capabilities for zero-shot GUI understanding without explicit RL training~\cite{yan2023gpt,zhang2023appagent}. However, these methods face distinct challenges: offline RL suffers from distribution shift when deployed to novel GUI configurations~\cite{levine2020offline}, while reward models struggle to adapt to interface changes that invalidate their training criteria~\cite{prudencio2023survey}. Even advanced VLMs like GPT-4o exhibit limitations in handling GUI complexity, as their lack of task-specific fine-tuning leads to misinterpretation of visual elements and interface dynamics~\cite{xie2024osworld,zhang2024ufo}. Recent hybrid approaches like~\cite{cheng2024seeclick} attempt to combine environment-free pre-training with targeted fine-tuning, but significant gaps remain in achieving robust generalization across diverse GUI ecosystems.

\begin{figure*}[htb]
    \centering
    \includegraphics[width=0.75\textwidth]{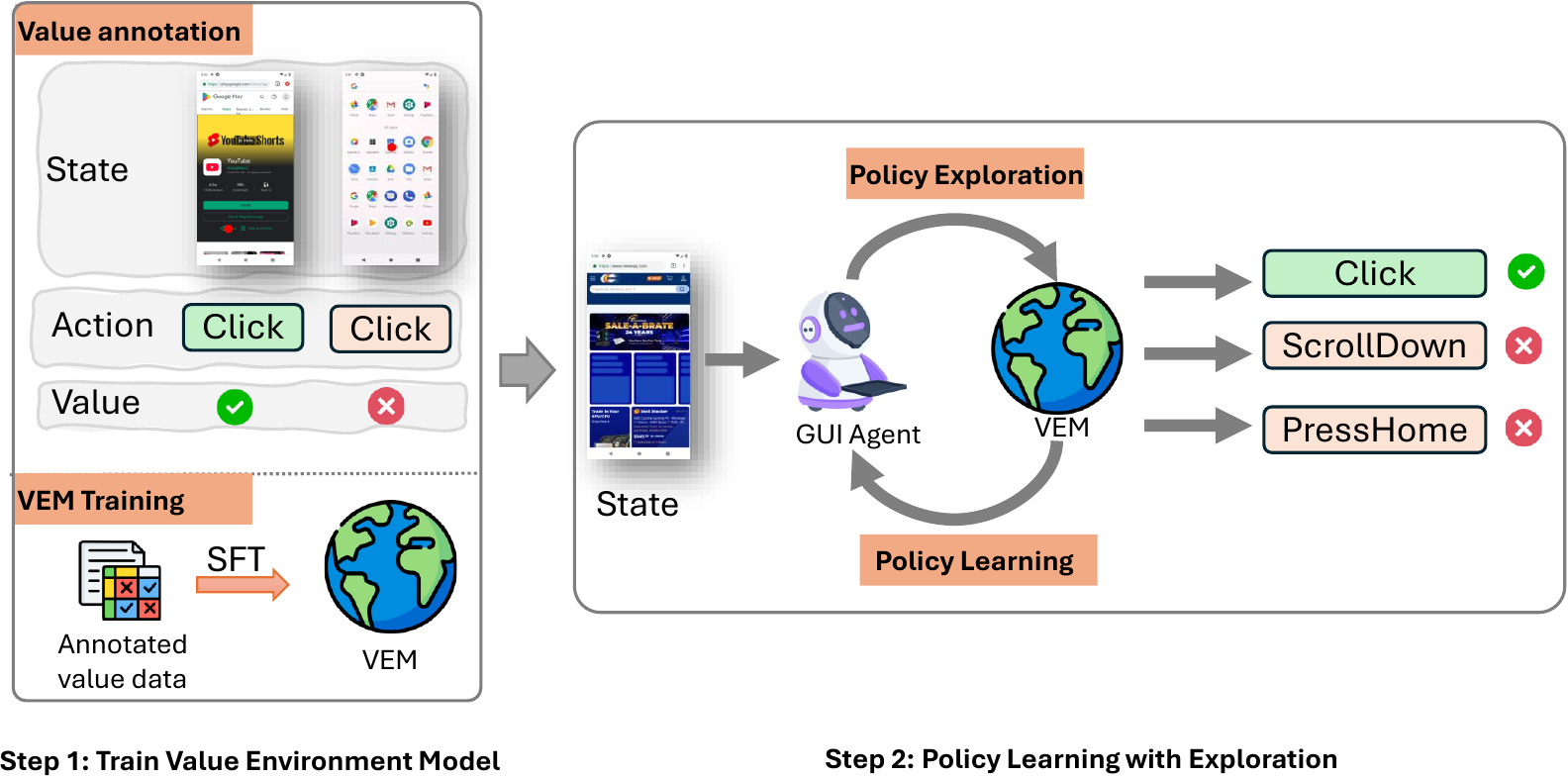}
    \caption{VEM Architecture: (1) Offline dataset annotation using GPT-4o's task understanding, and VEM training via supervised regression. (2) Policy optimization through frozen VEM maximization, encouraging the policy model to explore high-value actions.}
    \label{fig:structure}
    \vspace{-3mm}
\end{figure*}

\section{Method}
\label{sec:method_theory}

This section presents our method for training a GUI agent with offline data, followed by an extended theoretical analysis. As shown in Figure~\ref{fig:structure}, we first describe how to learn VEM from GPT-4o labeled data, then show that the resulting policy can achieve near-optimal performance under certain coverage and accuracy conditions. We further incorporate distribution-shift arguments to highlight the relationship between dataset quality and final policy performance.

\subsection{Preliminary}
\label{sec:setting}

We formalize GUI navigation as a Markov Decision Process $\mathcal{M} = (\mathcal{S}, \mathcal{A}, P, r, \gamma)$ where:
\begin{itemize}[leftmargin=*,nosep]
    \item States $s \in \mathcal{S}$: Task descriptions + interaction histories + current GUI screenshot.
    \item Actions $a \in \mathcal{A}$: \texttt{\{DUAL\_POINT, TYPE, PRESS\_BACK, SCROLL\_DOWN, ...\}} (see full actions in Appendix~\ref{appendix:actions}).
    \item $P(s'|s,a)$: Unknown environment dynamics.
    \item Rewards $r(s,a) \in \{0,1\}$: 0 for suboptimal choices, 1 for optimal actions.
    \item $\gamma \in [0,1)$: Discount factor.
\end{itemize}

Given an \textit{offline} dataset $\mathcal{D} = \{(s_i,a_i,r_i,s'_i)\}_{i=1}^N$ collected by unknown behavior policies or suboptimal behavior policy \(\beta\), our goal is to learn a policy $\pi_\phi(a|s)$ maximizing expected returns without environment interaction.

\subsection{Value Environment Model Training}
A core challenge in GUI automation is the scarcity of explicit reward signals that indicate whether a chosen action advances or hinders task completion. Our strategy is to generate coarse but direct supervision by leveraging GPT-4o’s understanding of the GUI context. Specifically, we assign each state-action pair a binary label, capturing whether the action is deemed beneficial or detrimental to the target task. This annotated supervision then guides the learning of a VEM, which is a state-action value function \(Q_\theta\), sidestepping the need for explicit environment interactions or hand-designed rewards.

\noindent\textbf{LLM-Guided Annotation.} 
For each state-action pair \((s, a)\) in our offline dataset \(\mathcal{D}\), we leverage GPT-4o with chain-of-thought reasoning \cite{wei2022chain} to generate binary labels \(\ell(s,a) \in \{0,1\}\), simulating human-like task progress assessment. These annotations provide coarse supervision signals, where \(\ell = 1\) indicates actions expected to aid task completion, while \(\ell = 0\) marks potentially counterproductive steps. This labeling scheme approximates long-term value through immediate assessments, circumventing the need for explicit reward engineering.

\noindent\textbf{Supervised Value Learning}.  
Using the annotated subset \(\widetilde{\mathcal{D}}=\{(s_i,a_i,\ell_i)\}\), we fine-tune a Qwen2VL~\cite{qwen2vl} to predict label values through mean squared error minimization:  
\[
\min_{\theta}\, \mathbb{E}_{(s,a,\ell)\sim\widetilde{\mathcal{D}}} \left[ \left(Q_\theta(s,a) - \ell\right)^2 \right]
\]  
The trained \(Q_\theta\) estimates action quality through environmental understanding distilled from GPT-4o's annotations, rather than online interactions, to reduce cost and latency during subsequent training.

\noindent\textbf{Stable Policy Guidance}.  
After convergence, we freeze \(Q_\theta\) as a fixed VEM to provide consistent action evaluations. While the binary labels represent simplified supervision, they effectively encode task progression patterns that guide subsequent policy learning. This approach maintains stability by decoupling environment modeling from policy optimization, while remaining fully offline-trainable.

\subsection{Policy Learning with the Frozen VEM}
Having established a VEM that can evaluate actions in any given GUI state, we now wish to derive a policy that selects actions maximizing the predicted value. By freezing \(Q_\theta\) as a fixed state-action value estimator, we transform policy learning into a stable optimization problem that leverages consistent value predictions without environment interactions. 

\noindent\textbf{Value Maximization Framework}.  
Our policy \(\pi_\phi(a \mid s)\) learns to select high-value actions by maximizing:
\[
\mathcal{J}(\pi_\phi) = \mathbb{E}_{s\sim \mathcal{D},\, a \sim \pi_\phi(\cdot \mid s)}\bigl[Q_\theta(s,a)\bigr],
\]
where the expectation is computed over states from the fixed dataset \(\mathcal{D}\) and actions from the learned policy. We implement this objective using Proximal Policy Optimization (PPO)~\cite{schulman2017proximal}, where each training iteration samples a mini-batch of states from \(\mathcal{D}\), generates candidate actions through \(\pi_\phi\), and updates \(\phi\) via gradient ascent on \(Q_\theta\)'s value estimates. This update rule increases the likelihood of actions that \(Q_\theta\) deems optimal, creating a feedback loop that aligns the policy with the VEM's understanding of action quality.

\noindent\textbf{Offline Training Advantages}.  
The frozen \(Q_\theta\) and static dataset \(\mathcal{D}\) enable purely offline learning, eliminating both the need for environment rollouts and the variance from on-policy data collection. This design ensures that policy updates rely solely on precomputed value estimates, avoiding the compounding errors that typically occur when alternating between policy improvements and value function updates. The resulting training process not only reduces computational costs but also stabilizes learning by maintaining fixed optimization targets throughout training.

This approach effectively converts policy optimization into a search over action selections that maximize a fixed quality metric, leveraging the VEM's environmental understanding while circumventing the exploration challenges inherent in GUI automation. The combination of value-driven updates and offline execution provides a practical solution to the sample efficiency and stability challenges in GUI policy learning.

\subsection{Theoretical Analysis}
We now provide a more advanced argument showing that if \(Q_\theta\) approximates \(Q^*\), i.e., the optimal value model, on the support of \(\mathcal{D}\), then the learned policy $\pi$ can achieve near-optimal returns. In addition, we introduce distribution shift considerations and demonstrate how coverage of \(\mathcal{D}\) influences policy quality.

\noindent\textbf{Offline Coverage and Value Approximation.} We introduce two conditions which bounds the suboptimality gap relative to the optimal policy \(\pi^*\):

\noindent\textit{Coverage Definition.}
For a policy \(\pi\), define the normalized state-action distribution \(d_\pi\) over \(\mathcal{D}\) (or the environment). Roughly speaking, \(d_\pi(s,a)\) captures how often \(\pi\) visits \((s,a)\). We say \(\pi\) is \emph{supported by} \(\mathcal{D}\) if, whenever \(\pi(a\mid s)>0\), the dataset \(\mathcal{D}\) contains sufficient samples of \((s,a)\). Let \(\|\pi-\beta\|\) measure how different \(\pi\) is from the behavior policy \(\beta\). If \(\|\pi-\beta\|\) is large, then \(\pi\) may choose actions not well-represented in \(\mathcal{D}\).

\noindent\textit{Approximate Q-Function.}
Assume there is a small \(\varepsilon\ge0\) such that
\[
\bigl|\,Q_\theta(s,a)-Q^*(s,a)\bigr|
\;\le\;\varepsilon
\quad
\forall\,(s,a)\in \mathcal{D}.
\]
This assumption states that on all \((s,a)\) pairs in \(\mathcal{D}\), \(Q_\theta\) is within \(\varepsilon\) of the optimal Q-value \(Q^*\).

\noindent\textbf{Performance Bound with Distribution Shift.} We introduce the extended performance bound theorem below:

\begin{theorem}[Extended Performance Bound]
\label{thm:performance_bound}
Let \(\widehat{\pi}\) maximizes \(\mathcal{J}(\pi) = \mathbb{E}_{s\sim\mathcal{D},\,a\sim\pi(\cdot\mid s)}[Q_\theta(s,a)]\). Under coverage and approximation conditions:
\begin{enumerate}[nosep,leftmargin=*]
    \item \textbf{Coverage}: \(\widehat{\pi}\) remains within the support of \(\mathcal{D}\). Formally, whenever \(\widehat{\pi}(a\mid s)>0\), \(\beta(a\mid s)>0\).
    \item \textbf{Approx. Q}: For all \((s,a)\in \mathcal{D}\), we have \(\bigl|Q_\theta(s,a) - Q^*(s,a)\bigr|\le\varepsilon\).
\end{enumerate}
Then there exists a constant \(c>0\) (depending on \(\gamma\) and the horizon) such that
\[
J(\pi^*) - J(\widehat{\pi})
\;\le\;
c\,\bigl(\varepsilon + \|\widehat{\pi}-\beta\|\bigr).
\]
Moreover, because \(\widehat{\pi}\) queries a \emph{fixed} Q-function from a static dataset, the variance of its gradient estimates can be significantly lower than that of an on-policy RL method interacting with a stochastic environment.
\end{theorem}

The proof details of Theorem~\ref{thm:performance_bound} are shown in Appendix~\ref{appendix:proof}.

\noindent\textbf{Implementation Consequences}.  
Theorem~\ref{thm:performance_bound} reveals two critical pathways for improving policy performance:

\begin{enumerate}
    \item \textbf{Enhancing Value Estimation}: \(\varepsilon\) measures how well \(Q_\theta\) matches \(Q^*\) on the offline data. It can be reduced \(\varepsilon\) through better VEM training (e.g., improved GPT-4o labeling or regression architecture)
    \item \textbf{Managing Distribution Shift}: \(\|\widehat{\pi}-\beta\|\) measures how far the learned policy strays from the dataset's distribution. It can be controlled via dataset diversification or policy constraints (e.g., collecting diverse or near-optimal transitions)
\end{enumerate}

\subsection{Stability Advantages}

Unlike on-policy RL, which repeatedly interacts with an environment and can suffer from high-variance or unsafe exploration, our offline approach trains solely from a fixed dataset and a frozen VEM:
\begin{itemize}
    \item \textbf{Fixed Surrogate Rewards:} Every query to \(Q_\theta\) for \((s,a)\) yields a deterministic estimate, eliminating environment stochasticity.
    \item \textbf{No On-Policy Data Collection:} The policy never modifies the dataset distribution; hence, distribution shift during training is avoided.
    \item \textbf{Reduced Gradient Variance:} Policy updates rely on consistent Q-values from \(Q_\theta\), typically leading to smoother convergence.
\end{itemize}
This results in a stable and efficient training process, especially beneficial for complex GUI tasks where real-time interactions may be expensive or infeasible.

Our method combines GPT-4o annotations with a VLM, e.g., Qwen2VL, to produce a frozen VEM \(Q_\theta\). A policy \(\pi_\phi\) then learns by maximizing \(\mathbb{E}[Q_\theta(s,a)]\) over states in the offline dataset \(\mathcal{D}\). The extended theoretical analysis reveals that if \(Q_\theta\) accurately approximates \(Q^*\) on the data support and the learned policy does not stray outside that support, then near-optimal performance can be guaranteed. Moreover, training stability improves by avoiding on-policy rollouts in a stochastic or expensive environment. We validate these insights experimentally, emphasizing how Q-function accuracy directly impacts final policy effectiveness. 

\section{Experiments}
In this section, we verify the performance of the model on two subsets of Android In-the-Wild (AITW)~\cite{zhang2024android} on the Android platform following DigiRL~\cite{bai2024digirl} in both offline and online settings.

\subsection{Data Collection}
\label{sec:data}

\paragraph{Training data for Critic Model.}
To align Qwen2VL with the input-output schema and evaluation framework of the critic model, we employed GPT-4o for multimodal data annotation and conducted systematic value assessments of agent actions. As depicted in Figure~\ref{fig:intro_case}, our hierarchical evaluation system defines two action quality levels based on task-oriented effectiveness.

\textbf{Level 1 (Suboptimal Actions).} manifest deviations from optimal task execution, specifically including: (1) erroneous text inputs compromising workflow integrity, (2) interface interactions triggering adversarial outcomes such as advertisement redirections, and (3) premature declarations of task completion prior to objective fulfillment. 

\textbf{Level 2 (Optimal Actions).} demonstrate maximally effective task-solving behaviors, characterized by three critical patterns: (1) verifiable task completion through interface state validation, (2) robust recovery strategies exemplified by \texttt{PRESS\_HOME} command execution for interface reset, and (3) context-aware selection of optimal entry points for subtask resolution.

The complete evaluation prompts and annotation protocols are formally specified in Appendix~\ref{appendix:prompt}.

\begin{table}[htb]
    \centering
        \caption{AITW dataset composition: training/testing episodes and interaction steps with action quality levels (1/2) across domains.}
    \adjustbox{max width=0.9\columnwidth}{
    \begin{tabular}{llcccc}
        \toprule
        Category & Split & Episodes & Steps & Level 1 & Level 2 \\
        \midrule
        \multirow{2}{*}{General} 
            & train & 436 & 3340 & 1187 & 2153 \\
            & test  & 100 & 777  & 214  & 563  \\
        \midrule
        \multirow{2}{*}{Webshopping} 
            & train & 560 & 6240 & 1939 & 4301 \\
            & test  & 91  & 772  & 273  & 499  \\
        \bottomrule
    \end{tabular}}
    \label{tab:dataset}
    \vspace{-3mm}
\end{table}

The AITW benchmark comprises 30K instructions with 715K operational trajectories. Adopting SeeClick~\cite{cheng2024seeclick} instruction-level partitioning, we analyze General and Webshopping domains. Table~\ref{tab:dataset} quantifies training/testing episodes and step distributions with action quality levels (Level 1/2). GPT-4o annotations achieve 90\% human consistency with 3-hour processing efficiency.

\paragraph{Data Format}
The reinforcement learning paradigm requires standardized data transformations across input modalities. For Qwen2VL, we normalize bounding box coordinates to device-agnostic [0,1] screen-space coordinates. Auto-GUI~\cite{zhang2023you} retains absolute coordinate pairs (touch/lift positions) for gesture modeling. Detailed action space configurations are provided in Section~\ref{sec:setting}.

\subsection{Implementation Details}
\label{sec:implementation}
\paragraph{Critic model} 
The Qwen2VL model underwent supervised fine-tuning through the LLaMAFactory framework to develop state classification capabilities. Our architectural modifications enabled hierarchical state prediction from multimodal inputs containing four distinct elements: textual task descriptions, historical interaction records, current screen imagery, and pending action sequences. The formal specification of input composition and prompting strategy appears in Appendix~\ref{appendix:prompt}.

Training leveraged distributed data parallelism on an 8-GPU NVIDIA A100 cluster, configured with 10 training epochs and a global batch size of 64 (8 samples per GPU). The optimization process employed AdamW with an initial learning rate of $1\times10^{-5}$, achieving convergence within 12 hours while maintaining computational efficiency through gradient accumulation strategies.

\paragraph{Policy model}
The policy architecture builds upon the Supervised Fine-Tuned (SFT) Auto-GUI foundation model while maintaining parameter freezing of the Critic module. Our implementation executed full-parameter optimization on a single NVIDIA A100 accelerator, configuring the training process with a batch size of 32 samples and $1\times10^{-6}$ across five training epochs. This configuration achieved stable convergence through progressive reward signal alignment, demonstrating parameter-efficient adaptation characteristics.

Notably, the gradient update strategy incorporated differential learning rate scheduling between the frozen Critic components and tunable policy layers, effectively balancing knowledge retention with operational flexibility. The complete training regimen required under 8 hours of computation time, reflecting optimized memory utilization patterns.

\paragraph{Baselines}
Our evaluation incorporates representative methods in embodied GUI interaction research. We consider GPT-4o~\cite{GPT-4o} as the foundational model without task-specific fine-tuning, alongside Auto-GUI~\cite{zhang2023you} trained through standard supervised fine-tuning (SFT). The comparison extends to specialized agents including CogAgent~\cite{hong2024cogagent} and SeeClick~\cite{cheng2024seeclick}, both employing SFT with expanded architectural configurations, as well as Digirl~\cite{bai2024digirl} which enhances Auto-GUI's capabilities through a combined offline pretraining and online reinforcement learning framework. This baseline selection captures diverse approaches to GUI interaction while maintaining methodological coherence across different training paradigms and architectural developments.

\subsection{VEM Performance}  
We evaluate the performance of our VEM models on both the General and WebShopping datasets, as shown in Table~\ref{tab:vem}. Our trained VEM achieves an F1 score of 78\% and an accuracy of 69\% on the two datasets, demonstrating high performance reliability. This level of accuracy is sufficient to drive policy optimization via exploration. 

\begin{figure*}[t]
    \centering
    \includegraphics[width=0.9\textwidth]{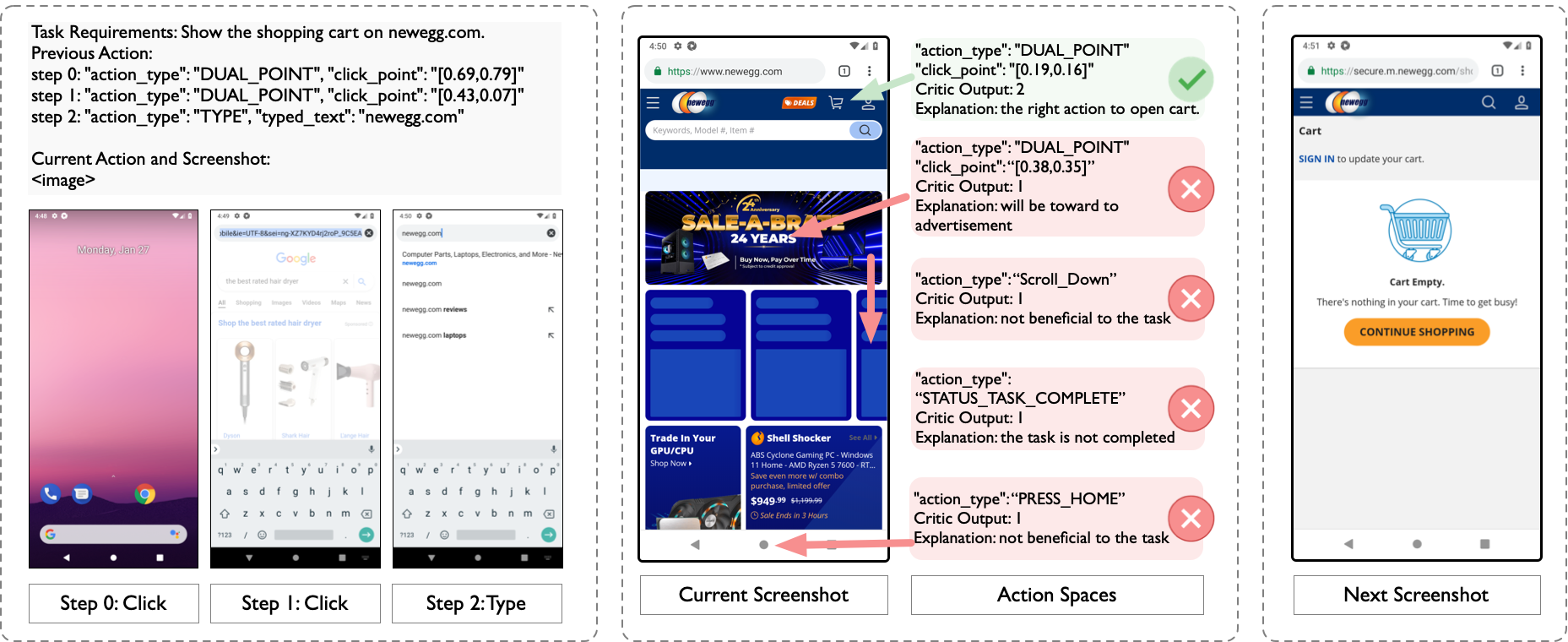}
    \vspace{-0.5em}
    \caption{Action space exploration patterns demonstrated by the value model during policy training.}
    \vspace{-0.5em}
    \label{fig:action}
    \vspace{-2mm}
\end{figure*}

As shown in a case in Figure~\ref{fig:action}, the exploration trajectory exhibits three distinct phases: an initial conservative exploitation phase with limited action diversity, followed by a transitional exploration phase characterized by increasing behavioral variance, and finally, an expanded exploration phase that systematically navigates high-dimensional action spaces. This phased progression empirically demonstrates the model's ability to autonomously balance the exploitation-exploration tradeoff while maintaining stable policy improvement gradients.

\begin{figure}[t]
    \centering
    \includegraphics[width=0.45\textwidth]{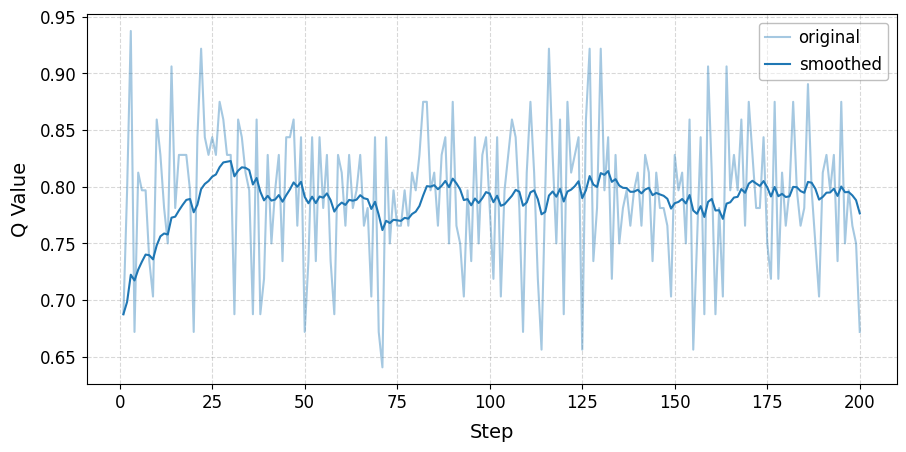}
    \vspace{-1em}
    \caption{Q-value loss progression during policy model training.}
    \label{fig:loss}
    \vspace{-3mm}
\end{figure}

\subsection{Main Results}

\begin{table}[t]
\caption{Performance on the VEM.}
\label{tab:vem}
\resizebox{\columnwidth}{!}{
\begin{tabular}{c|c|c|c|c}
\hline
\textbf{Dataset}     & \textbf{Precision} & \textbf{Recall} & \textbf{F1} & \textbf{Accuracy} \\ \hline
\textbf{General} & 0.81 & 0.78            & 0.79        & 0.71              \\ \hline
\textbf{Webshopping} & 0.82                & 0.79            & 0.80        & 0.75              \\ \hline
\end{tabular}
}
\end{table}

\label{sec:main_result}
We employ two complementary evaluation schemes: (1) \textbf{Offline Evaluation} computes step/task success rates (SRs) by comparing predicted actions with human annotations in test data; (2) \textbf{Online Evaluation} deploys the agent in real Android environments (aligned with DigiRL's setup) using GPT-4o as automated judge, with 10-step attempt limits and duplicate task removal.

The offline setting measures precise action matching through deterministic verification, while the online test evaluates practical deployment capability under environmental dynamics. This dual approach separates basic competence assessment from real-world performance analysis, ensuring comprehensive model validation.

\begin{figure*}[t] 
    \centering
    \includegraphics[width=1\textwidth]{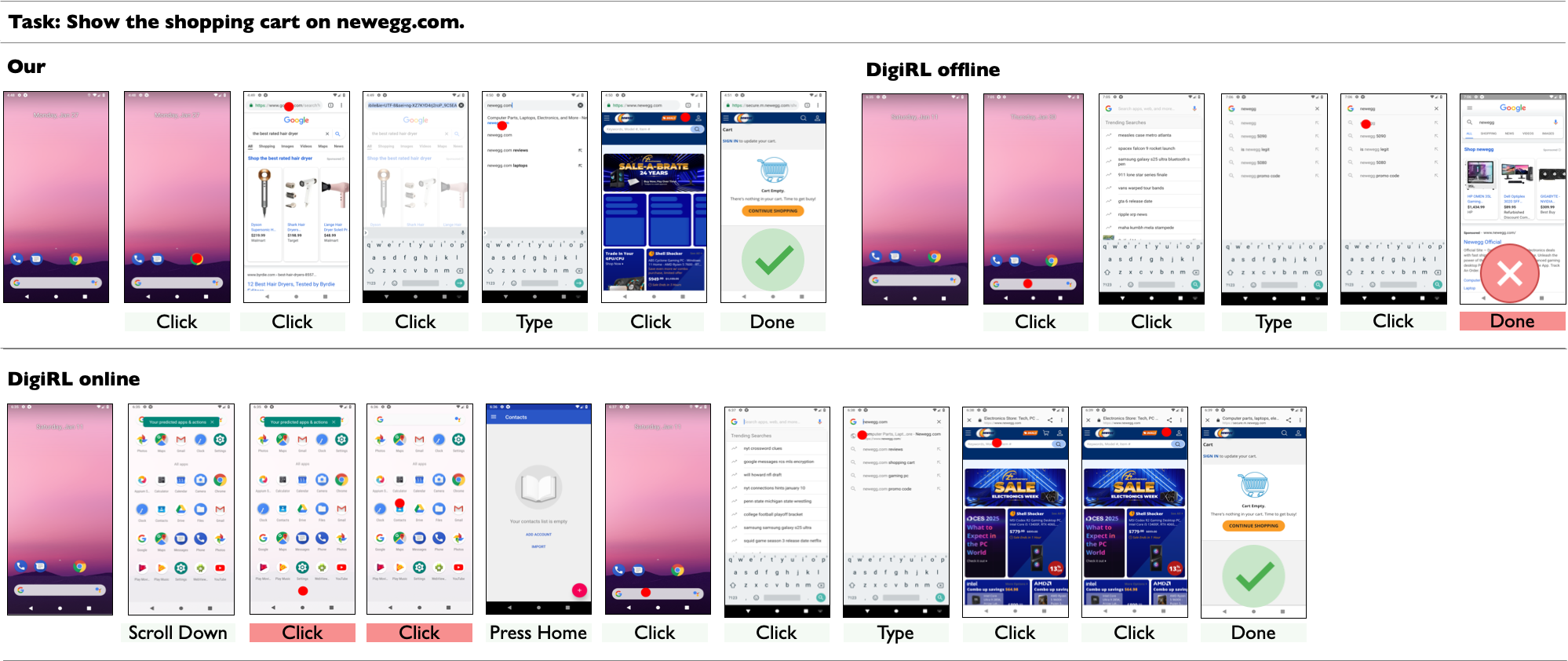}
    \vspace{-2em}
    \caption{A case study of task execution trajectory comparison with DigiRL.}
    \label{fig:main_case}
    \vspace{-2mm}
\end{figure*}

{\small
\begin{table}[t]
\centering
\caption{Offline evaluation results on the AITW benchmark.}
\resizebox{\columnwidth}{!}{ 
\begin{tabular}{lcccc}
    \toprule
    & \multicolumn{2}{c}{General} & \multicolumn{2}{c}{Webshopping} \\
    \cmidrule(lr){2-3} \cmidrule(lr){4-5}
    & {Step SR} & {Task SR} & {Step SR} & {Task SR} \\
    \midrule
    GPT-4o & 32.0 & 9.0 & 30.6 & 0.0 \\
    Auto-GUI & 83.3 & 20.0 & 78.0 & 18.0 \\
    CogAgent & 73.7 & 16.0 & 72.2 & 9.0 \\
    Seeclick & 83.3 & 26.0 & 79.1 & 18.0 \\
    Digirl (offline) & 80.0 & 21.0 & 80.2 & 14.0 \\
    Digirl (online) & \textbf{83.3} & 23.0 & 78.5 & 17.0 \\
    Ours & 82.6 & \textbf{28.0} & \textbf{81.1} & \textbf{21.0} \\
    \bottomrule
\end{tabular}
}
\label{tab:offline_result}
\end{table}
}

The experimental results shown in Table~\ref{tab:offline_result} demonstrate significant advantages of our approach. With the same Auto-GUI policy model, our method achieves 28.0\% task success rate on General domain and 21.0\% on Webshopping, surpassing all baselines including DigiRL's online variant (23.0\% and 17.0\% respectively). This 21.7\% relative improvement over the strongest baseline confirms that value model supervision effectively guides policy optimization without requiring additional data collection or environmental interaction.

Notably, our approach outperforms DigiRL's online training results despite operating purely offline, particularly in task-level metrics where conventional RL methods typically struggle. The complete failure of GPT-4o on Webshopping tasks (0\% success rate) further emphasizes the necessity of domain-specific training, which our specialized models address through tailored value estimation. These findings establish that structured credit assignment can compensate for the absence of online exploration, achieving state-of-the-art performance through more sample-efficient learning.

{\small
\begin{table}[t]
\centering
\caption{Online deployment performance on AITW benchmark.}
\resizebox{\columnwidth}{!}{ 
\begin{tabular}{lcccc}
    \toprule
    & \multicolumn{2}{c}{General} & \multicolumn{2}{c}{Webshopping} \\
    \cmidrule(lr){2-3} \cmidrule(lr){4-5}
    & {Task SR} & {Step Length} & {Task SR} & {Step Length} \\
    \midrule
    GPT-4o & 0.00 & 9.00 & 0.00 & 9.91 \\
    Auto-GUI & 28.81 & 7.92 & 2.86 & 9.34 \\
    CogAgent & 38.98 & 7.23 & 14.29 & 7.80 \\
    Seeclick & 25.42 & 8.80 & 11.43 & 9.80 \\
    Digirl (offline) & 38.98 & 7.61 & 14.29 & 8.26 \\
    Digirl (online) & 38.98 & 7.25 & 11.43 & 7.37 \\
    Ours & \textbf{42.37} & 7.83 & \textbf{14.29} & 7.86 \\
    \bottomrule
\end{tabular}
}
\label{tab:online_result}
\vspace{-3mm}
\end{table}
}

The online evaluation demonstrates our method's robustness in dynamic environments. Under real-world conditions with random disturbances (e.g., advertisement popups), our approach achieves superior task success rates (42.37\% General, 14.29\% Webshopping), outperforming all baselines including DigiRL's online variant (38.98\%, 11.43\%). This 8.7\% improvement on General tasks highlights enhanced generalization capability despite equivalent environmental access.

Analysis of interaction patterns reveals distinct behavioral characteristics: While supervised learning methods (Auto-GUI:7.92 steps, CogAgent:7.23) exhibit moderate trajectory lengths, our RL-enhanced framework maintains near-optimal path efficiency (7.83 General, 7.86 Webshopping) without sacrificing success rates. This contrasts with DigiRL's online strategy that prioritizes minimal steps (7.25 General) at the cost of reduced accuracy, suggesting our method better balances exploration and exploitation. More cases can be founded in Appendix~\ref{appendix:case}.

The performance discrepancy between DigiRL's offline (14.29\% Webshopping) and online (11.43\%) results stems from overfitting to simplified training scenarios, whereas our approach's adaptive credit assignment mechanism proves more resilient to environmental novelty.

Our evaluation demonstrates that with limited training data, our method consistently outperforms DigiRL's offline approach and matches its online variant, validating the sample efficiency of our framework.

\subsection{Analysis}
\label{sec:analysis}

\paragraph{Training stability}

As evidenced in Figure~\ref{fig:loss}, our policy training methodology achieves substantially improved stability compared to standard Proximal Policy Optimization (PPO) frameworks. While typical PPO implementations exhibit erratic loss fluctuations due to gradient update instability and high-variance advantage estimation, our approach mitigates these instability patterns through gradient-constrained policy updates and adaptive exploration regulation. The stabilized training dynamics not only reduce convergence variance but also enhance both training efficiency and model generalizability, addressing fundamental challenges in policy optimization processes.

\paragraph{Case Study}  
Deploying GUI agents in real-world scenarios presents inherent challenges due to real-time system dynamics and environmental stochasticity, often leading to unintended navigation to advertisement interfaces during task execution. While models trained solely via SFT exhibit limited error recovery in such open-world settings, our value-guided approach enables robust policy adaptation.  

As shown in Figure~\ref{fig:main_case}, our method achieves minimal-path task completion through precise state valuation, whereas DigiRL offline baselines frequently misidentify terminal states, and DigiRL online methods suffer from suboptimal trajectories with recovery loops. This capability arises from the value model’s continuous interpretation of environmental feedback, facilitating dynamic policy correction absent in conventional training paradigms. Extended comparative analyses with alternative baselines are provided in Appendix~\ref{appendix:case}.

\section{Conclusion}
We presents an environment-free RL framework for GUI automation that decouples value estimation from policy optimization through a Value Environment Model (VEM). Our approach replaces error-prone next-state simulations with semantic reasoning over GUI elements, enabled by offline learning from human demonstration data. The two-stage training paradigm achieves structured credit assignment without environmental interaction, while maintaining procedural efficiency comparable to environment-based methods. Experimental results on Android-in-the-Wild demonstrate superior task success rates over existing environment-free approaches and significant improvements in generalization capability compared to vision-language models. The framework establishes semantic-driven value estimation as an effective pathway for layout-agnostic GUI automation with sample efficiency. In the future, we plan to explore self-supervised approaches for training the value model, aiming to reduce labeling overhead and further improve scalability.

\bibliography{ref}

\newpage
\appendix
\onecolumn
\section*{Appendix}

\section{Actions}
\label{appendix:actions}
The available actions include DUAL\_POINT, TYPE, PRESS\_BACK, PRESS\_HOME, SCROLL\_DOWN, SCROLL\_UP, SCROLL\_LEFT, SCROLL\_RIGHT, PRESS\_ENTER, STATUS\_TASK\_COMPLETE, and STATUS\_TASK\_IMPOSSIBLE.

\section{Proof of Extended Performance Bound}
\label{appendix:proof}

\begin{proof}[Proof]
First, by the approximate accuracy assumption, \(Q_\theta\approx Q^*\) within \(\varepsilon\) on \(\mathcal{D}\). Thus, for any \((s,a)\) in the dataset support, maximizing \(Q_\theta(s,a)\) is nearly the same as maximizing \(Q^*(s,a)\). Second, the coverage requirement ensures that \(\widehat{\pi}\) does not select actions absent from \(\mathcal{D}\). Standard distribution shift arguments (\emph{cf.} \cite{levine2020offline}) show that if \(\|\widehat{\pi}-\beta\|\) is small, then \(\widehat{\pi}\) does not deviate far from the offline distribution. Combining these points yields a bound
\[
J(\pi^*) - J(\widehat{\pi})
\;\le\; 
c(\varepsilon + \|\widehat{\pi}-\beta\|)
\]
for some constant \(c\). Finally, since all policy updates use \(\bigl\{(s,a)\sim\mathcal{D}\bigr\}\) and a fixed Q-model, the training variance is typically much lower than that of an on-policy method that must repeatedly sample from a potentially noisy environment.
\end{proof}

\section{Prompt}
\label{appendix:prompt}

\textbf{Prompt of GPT-4o input}
\begin{tcolorbox}[colback=white,colframe=black,boxrule=1pt,width=\textwidth,arc=0pt,breakable]
\begin{lstlisting}
As an expert in the field of GUI and reinforcement learning, you will receive complete screenshots and textual descriptions of interactions for a given task. You need to evaluate a specific step in terms of its value within the task chain, similar to what a value function does in reinforcement learning. Detailed criteria and standards are given below.

## Explanation of the input content:
1. Task: Brief description of the current GUI task, such as implementing the "Get Hong Kong hotel prices" task in Android GUI.
2. Complete operation description and corresponding screenshot sequence for the task
   (1) Text description of operations: Contains 11 types of GUI operations. Specific fields and their meanings are as follows:
      [1] DUAL_POINT: Double-click on a specific position on the screen. If it is a link or software, it will enter; if it is text, it will be selected. The "click_point" is represented by a two-dimensional array indicating the position of the click, relative to the top-left corner of the screenshot and within a range from 0.0 to 1.0.
         - example: "action_type": "DUAL_POINT", "click_point": [0.5, 0.5]
      [2] TYPE: An action type that sends text. Note that this simply sends text and does not perform any clicks for element focus or enter presses for submitting text.
         - example: "action_type": "TYPE", "typed_text": "capital of England"
      [3] PRESS_BACK: Return to the previous page. Usually the previous webpage.
         - example: "action_type": "PRESS_BACK"
      [4] PRESS_HOME: Return to the system home page. Use this action to return to the home screen when the current screen is not the desired one, so you can reselect the program you need to enter.
         - example: "action_type": "PRESS_HOME"
      [5] PRESS_ENTER: Press the enter key to execute a step. Generally, after confirming the input text, use this action to start the search.
         - example: "action_type": "PRESS_ENTER"
      [6] STATUS_TASK_COMPLETE: An action used to indicate that the desired task has been completed and resets the environment. This action should also be used if the task is already completed and there is nothing more to do. For example, the task is to turn on the Wi-Fi when it is already on.
         - example: "action_type": "STATUS_TASK_COMPLETE"
      [7] STATUS_TASK_IMPOSSIBLE: An action used to indicate that the desired task is impossible to complete and resets the environment. This can result from various reasons including UI changes, Android version differences, etc.
         - example: "action_type": "STATUS_TASK_IMPOSSIBLE"
      [8] SCROLL_DOWN: Scroll down.
         - example: "action_type": "SCROLL_DOWN"
      [9] SCROLL_UP: Scroll up.
         - example: "action_type": "SCROLL_UP"
      [10] SCROLL_LEFT: Scroll left.
         - example: "action_type": "SCROLL_LEFT"
      [11] SCROLL_RIGHT: Scroll right.
         - example: "action_type": "SCROLL_RIGHT"
   (2) Corresponding screenshot before each operation. If the operation is of the "DUAL_POINT" type, the click position is marked with a red dot in the image.    
3. The current action to be evaluated and the corresponding screenshot.

## Evaluation Criteria:
Here are the detailed descriptions of the two levels. Attention needs to be paid to whether the action taken based on the current screenshot promotes efficient task execution, rather than the relevance of the content shown in the current screenshot to the task:
   Level 1: The action is not the optimal choice for completing the task at this moment, which may lead to deviations from the task flow. For example:
      (1) Incorrect text input.
      (2) Clicking a button that might lead to an advertisement.
      (3) Announcing the task's success when it has not actually been achieved.
   Level 2: The action is the optimal and correct choice for completing the task at this moment. For example:
      (1) When showing task completion, the displayed content can fully achieve it.
      (2) When entering an unrelated interface, you can return to the main screen by executing "PRESS_HOME."
      (3) Selecting the most correct entry point to complete the current task.

## Output requirements:
- Format: {"rating": int, "explanation": str}. Do not include any additional characters beyond this format
- The "rating" field should be represented by the number 1 or 2 indicating the evaluation level. The "explanation" field should explain the evaluation process that led to this rating, without including descriptions of operations after the current step (future operations are considered unknown).

## Example Input:
Task Requirements: What is the capital of England?
Action and ScreenShot:
step 0: "action_type": "DUAL_POINT", "click_point": "[0.524, 0.06]"
step 1: "action_type": "TYPE", "typed_text": "capital of England"
step 2: "action_type": "PRESS_ENTER"
step 3: "action_type": "STATUS_TASK_COMPLETE"
Current Action:
step 2: "action_type": "PRESS_ENTER"

## Example Output:
{"rating": 2, "explanation": "The action of pressing enter after typing 'capital of England' is an appropriate step to get the answer to the task requirement of finding out the capital of England, which is an optimal action towards achieving the task goal."}

Task Requirements: {}
Action and ScreenShot: {}
Current Action: 
{}
\end{lstlisting}
\end{tcolorbox}

\textbf{Prompt of critic input}
\begin{tcolorbox}[colback=white,colframe=black,boxrule=1pt,width=\textwidth,arc=0pt,breakable]
\begin{lstlisting}
As an expert in the field of GUI and reinforcement learning, you will receive textual descriptions of history interactions for a given task. You need to evaluate the current action, similar to what a value function does in reinforcement learning. Detailed criteria and standards are given below.

## Explanation of the input content:
1. Task: Brief description of the current GUI task, such as implementing the "Get Hong Kong hotel prices" task in Android GUI.
2. Description of History operation
   Contains 11 types of GUI operations. Specific fields and their meanings are as follows:
   [1] DUAL_POINT: Double-click on a specific position on the screen. If it is a link or software, it will enter; if it is text, it will be selected. The "click_point" is represented by a two-dimensional array indicating the position of the click, relative to the top-left corner of the screenshot and within a range from 0.0 to 1.0.
      - example: "action_type": "DUAL_POINT", "click_point": [0.5, 0.5]
   [2] TYPE: An action type that sends text. Note that this simply sends text and does not perform any clicks for element focus or enter presses for submitting text.
      - example: "action_type": "TYPE", "typed_text": "capital of England"
   [3] PRESS_BACK: Return to the previous page. Usually the previous webpage.
      - example: "action_type": "PRESS_BACK"
   [4] PRESS_HOME: Return to the system home page. Use this action to return to the home screen when the current screen is not the desired one, so you can reselect the program you need to enter.
      - example: "action_type": "PRESS_HOME"
   [5] PRESS_ENTER: Press the enter key to execute a step. Generally, after confirming the input text, use this action to start the search.
      - example: "action_type": "PRESS_ENTER"
   [6] STATUS_TASK_COMPLETE: An action used to indicate that the desired task has been completed and resets the environment. This action should also be used if the task is already completed and there is nothing more to do. For example, the task is to turn on the Wi-Fi when it is already on.
      - example: "action_type": "STATUS_TASK_COMPLETE"
   [7] STATUS_TASK_IMPOSSIBLE: An action used to indicate that the desired task is impossible to complete and resets the environment. This can result from various reasons including UI changes, Android version differences, etc.
      - example: "action_type": "STATUS_TASK_IMPOSSIBLE"
   [8] SCROLL_DOWN: Scroll down.
      - example: "action_type": "SCROLL_DOWN"
   [9] SCROLL_UP: Scroll up.
      - example: "action_type": "SCROLL_UP"
   [10] SCROLL_LEFT: Scroll left.
      - example: "action_type": "SCROLL_LEFT"
   [11] SCROLL_RIGHT: Scroll right.
      - example: "action_type": "SCROLL_RIGHT"
3. The current action to be evaluated and the corresponding screenshot(the screenshot before each operation. If the operation is of the "DUAL_POINT" type, the click position is marked with a red dot in the image.)

## Evaluation Criteria:
Here are the detailed descriptions of the two levels. Attention needs to be paid to whether the action taken based on the current screenshot promotes efficient task execution, rather than the relevance of the content shown in the current screenshot to the task:
   Level 1: The action is not the optimal choice for completing the task at this moment, which may lead to deviations from the task flow. For example:
      (1) Incorrect text input.
      (2) Clicking a button that might lead to an advertisement.
      (3) Announcing the task's success when it has not actually been achieved.
   Level 2: The action is the optimal and correct choice for completing the task at this moment. For example:
      (1) When showing task completion, the displayed content can fully achieve it.
      (2) When entering an unrelated interface, you can return to the main screen by executing "PRESS_HOME."
      (3) Selecting the most correct entry point to complete the current task.

## Output requirements: 1 or 2 (INT)

## Example Input:
Task Requirements: What is the capital of England?
Previous Action:
step 0: "action_type": "DUAL_POINT", "click_point": "[0.524, 0.06]"
step 1: "action_type": "TYPE", "typed_text": "capital of England"
Current Action and Screenshot:
step 2: "action_type": "PRESS_ENTER"

## Example Output:
2

Task Requirements: {}
Previous Action: 
{}
Current Action and Screenshot: 
<image>
{}
\end{lstlisting}
\end{tcolorbox}

\section{Case Study}
\label{appendix:case}

\begin{figure*}[htb]
    \centering
    \includegraphics[width=1.0\textwidth]{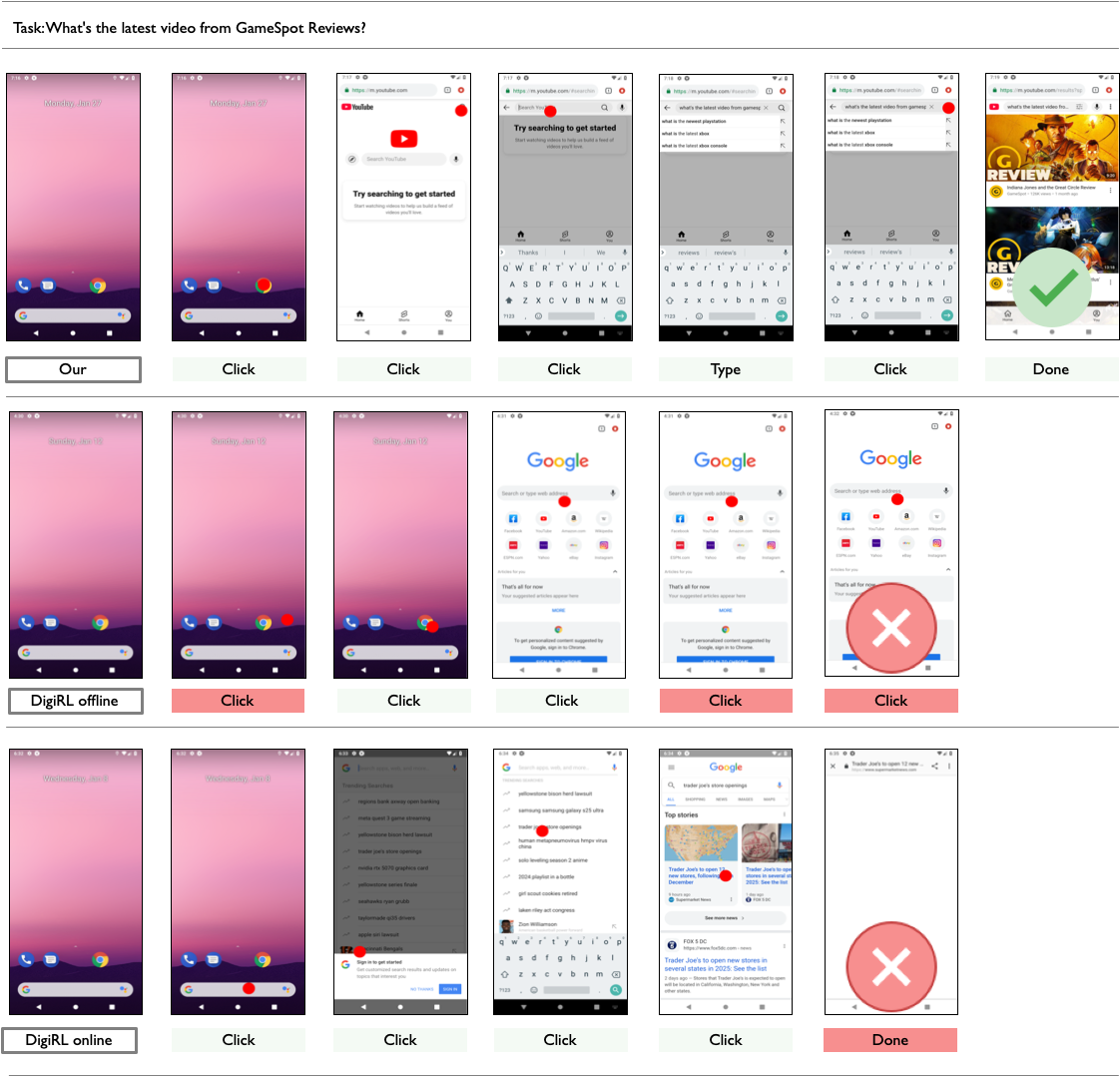}
    \caption{Case study of Ours, DigiRL offline and DigiRL online on the task: what's the latest video from GameSpot reviews?}
    \label{fig:case1}
\end{figure*}

\begin{figure*}[htb]
    \centering
    \includegraphics[width=1.0\textwidth]{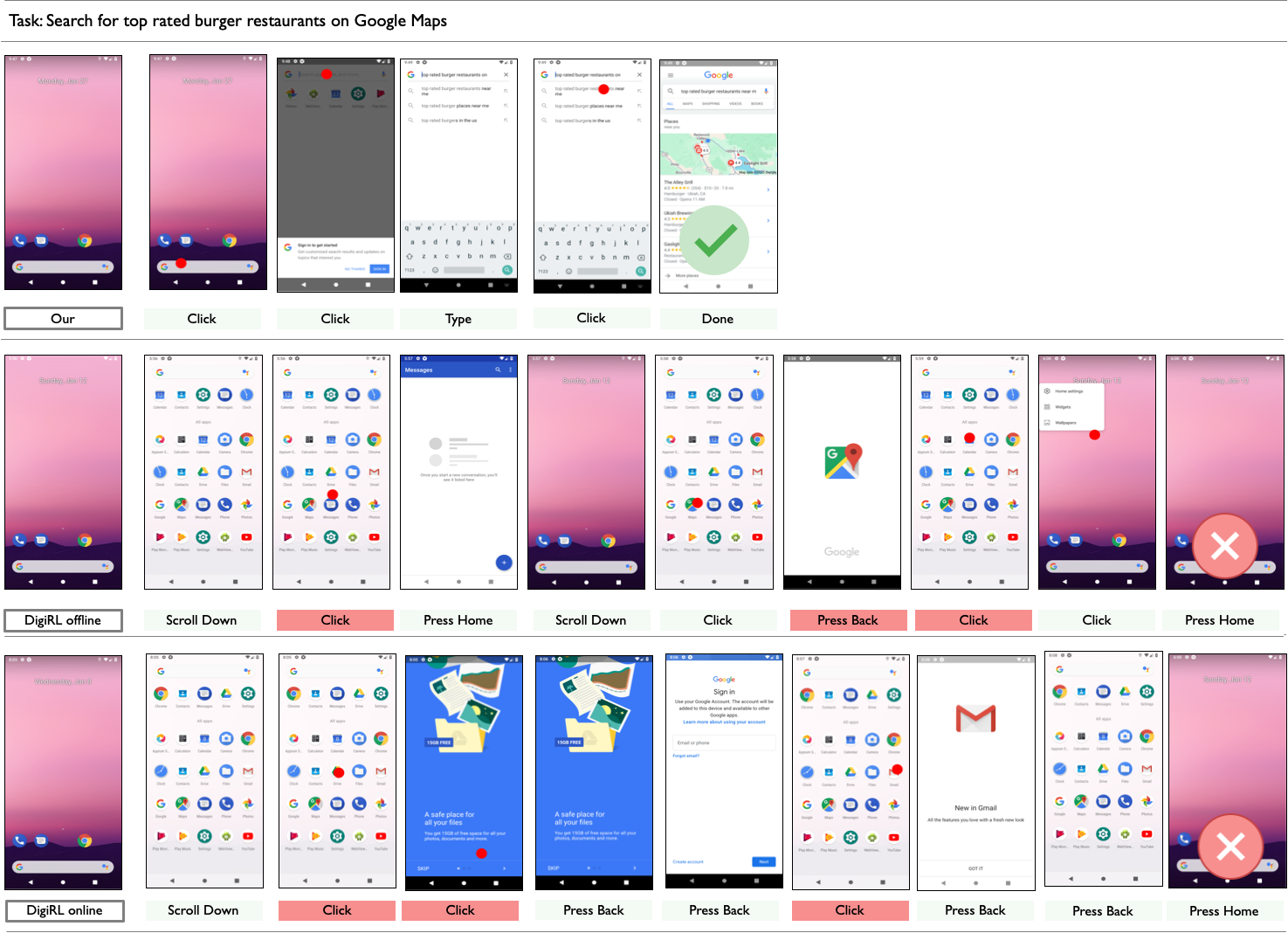}
    \caption{Case study of Ours, DigiRL offline and DigiRL online on the task: search for top rated burger restaurants on Google Maps.}
    \label{fig:case2}
\end{figure*}

\begin{figure*}[htb]
    \centering
    \includegraphics[width=1.0\textwidth]{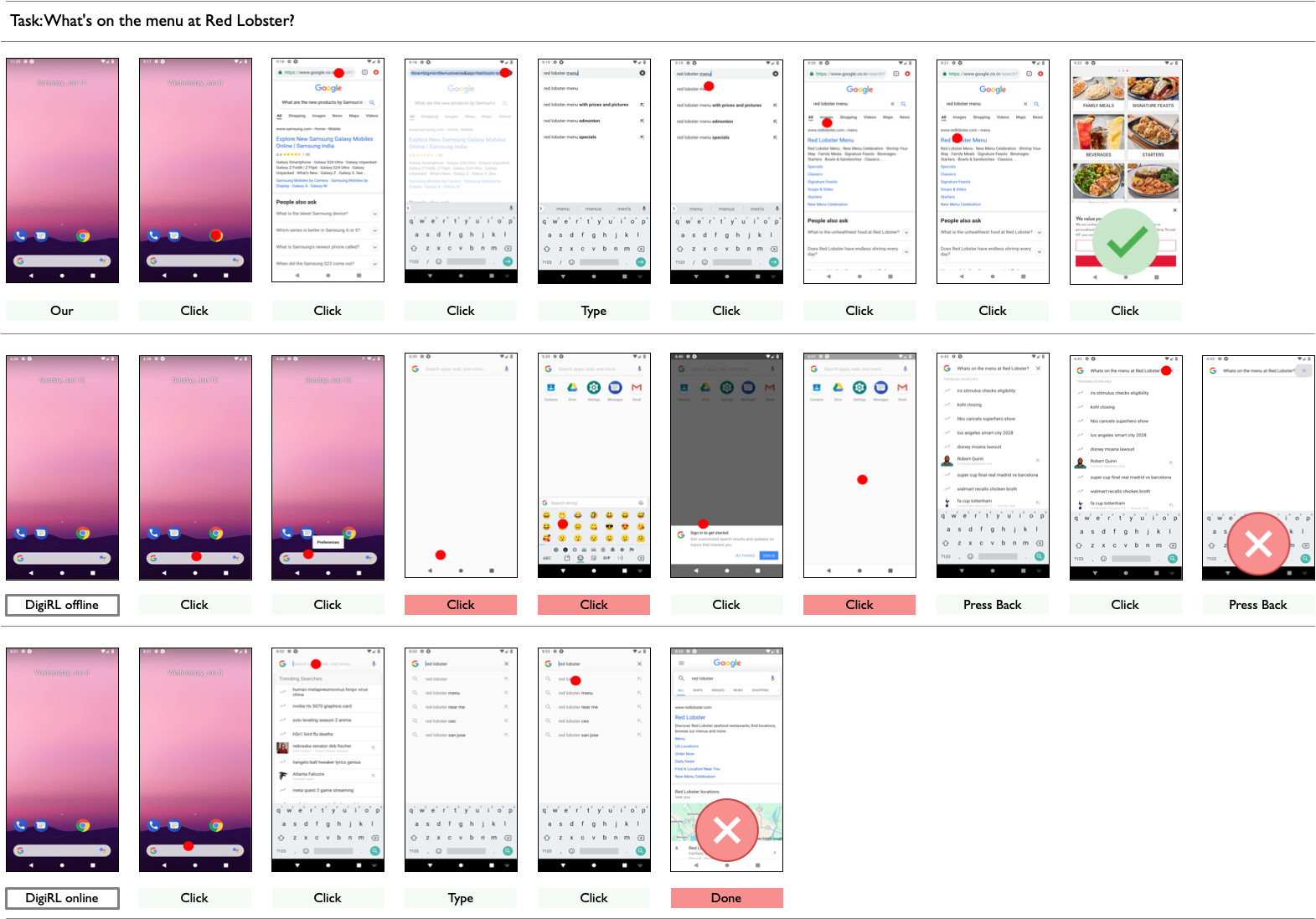}
    \caption{Case study of Ours, DigiRL offline and DigiRL online on the task: what's on the menu at Red Lobster?}
    \label{fig:case3}
\end{figure*}

\begin{figure*}[htb]
    \centering
    \includegraphics[width=1.0\textwidth]{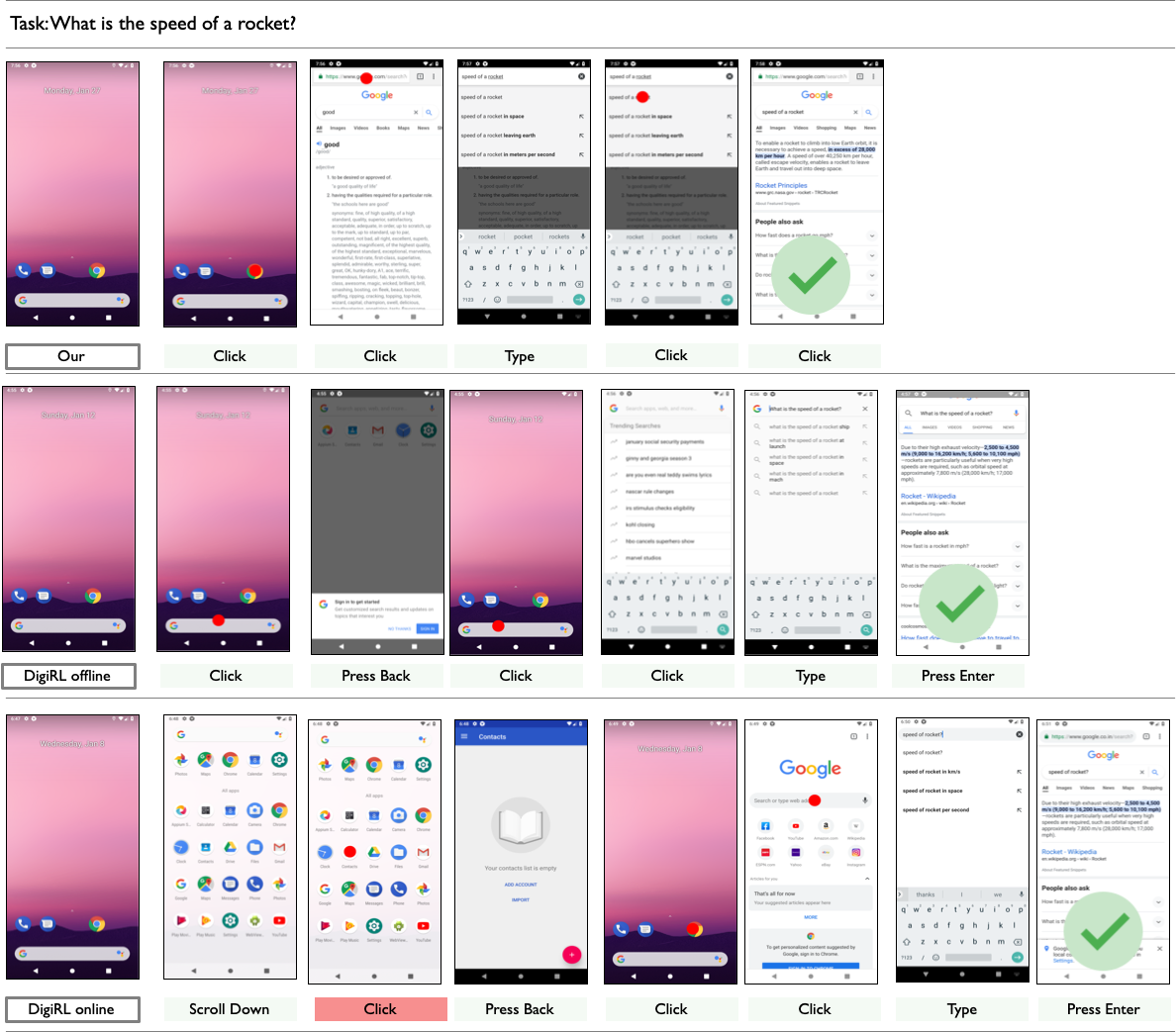}
    \caption{Case study of Ours, DigiRL offline and DigiRL online on the task: what is the speed of a rocket?}
    \label{fig:case4}
\end{figure*}

Here we randomly sample cases (Figure~\ref{fig:case1}, \ref{fig:case2}, \ref{fig:case3}, \ref{fig:case4}), from our test experiments to show the difference between our method and baselines.

\end{document}